\documentclass[letterpaper]{article}
\usepackage{uai2018}
\usepackage[margin=1in]{geometry}

\usepackage{hyperref}
\usepackage{url}
\usepackage{amsmath}
\usepackage{amsfonts}
\usepackage{caption}
\usepackage[normalem]{ulem}
\usepackage{booktabs}
\usepackage{graphicx} 
\usepackage{float}
\graphicspath{{./images/}}
\usepackage{lipsum}
\usepackage{color}

\usepackage[ruled]{algorithm2e}
\usepackage{algorithmic}

\usepackage{natbib}

\makeatletter
\newcommand{\subalign}[1]{%
  \vcenter{%
    \Let@ \restore@math@cr \default@tag
    \baselineskip\fontdimen10 \scriptfont\tw@
    \advance\baselineskip\fontdimen12 \scriptfont\tw@
    \lineskip\thr@@\fontdimen8 \scriptfont\thr@@
    \lineskiplimit\lineskip
    \ialign{\hfil$\m@th\scriptstyle##$&$\m@th\scriptstyle{}##$\crcr
      #1\crcr
    }%
  }
}
\makeatother

\newcommand\blfootnote[1]{%
  \begingroup
  \renewcommand\thefootnote{}\footnote{#1}%
  \addtocounter{footnote}{-1}%
  \endgroup
}

\usepackage{etoolbox}
\makeatletter
\patchcmd{\@lbibitem}
  {]}
  {][\theNAT@ctr] }
  {}{}
\makeatother

\usepackage{times}

\title{The Variational Homoencoder:\\ Learning to learn high capacity generative models from few examples}

\author{
\textbf{Luke B. Hewitt}\textsuperscript{12} \qquad 
\textbf{Maxwell I. Nye}\textsuperscript{1} \qquad 
\textbf{Andreea Gane}\textsuperscript{1} \qquad 
\textbf{Tommi Jaakkola}\textsuperscript{1} \qquad 
\textbf{Joshua B. Tenenbaum}\textsuperscript{1}   \vspace{1em}\\
\textsuperscript{1}Massachusetts Institute of Technology \\
\textsuperscript{2}MIT-IBM Watson AI Lab
}

\begin{document}

\maketitle

\begin{abstract}
Hierarchical Bayesian methods can unify many related tasks (e.g. $k$-shot classification, conditional and unconditional generation) as inference within a single generative model. However, when this generative model is expressed as a powerful neural network such as a PixelCNN, we show that existing learning techniques typically fail to effectively use latent variables. To address this, we develop a modification of the Variational Autoencoder in which encoded observations are decoded to new elements from the same class. This technique, which we call a \textit{Variational Homoencoder} (VHE), produces a hierarchical latent variable model which better utilises latent variables. We use the VHE framework to learn a hierarchical PixelCNN on the Omniglot dataset, which outperforms all existing models on test set likelihood and achieves strong performance on one-shot generation and classification tasks. We additionally validate the VHE on natural images from the YouTube Faces database. Finally, we develop extensions of the model that apply to richer dataset structures such as factorial and hierarchical categories.


\end{abstract}

\section{INTRODUCTION}
\blfootnote{A PyTorch implementation of the Variational Homoencoder can be found at \href{http://github.com/insperatum/vhe}{github.com/insperatum/vhe}. Supplement can be found in \href{http://auai.org/uai2018/proceedings/supplements/Supplementary-Paper351.pdf}{the UAI 2018 proceedings}.}Learning from few examples is possible only with strong inductive biases. In machine learning these biases can be hand designed, such as a model's parametrisation, or can be the result of a meta-learning algorithm. Furthermore they may be task-specific, as in discriminative modelling, or may describe the world causally so as to be naturally reused across many tasks. Recent work has approached one- and few-shot learning from all of these perspectives. 

\begin{figure*}[t]
\label{fig:headline}
\center
\includegraphics[width=0.9\textwidth]{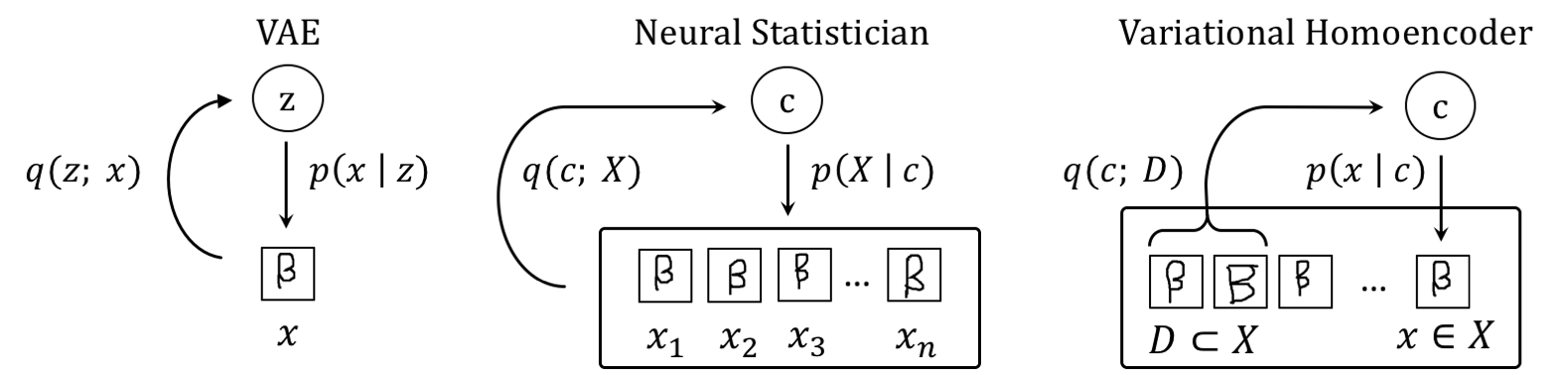}
\caption{Single step of gradient training in various models. A VAE treats all datapoints as independent, so only a single random element need be encoded (with $q(z;x)$) and decoded (with $p(x|z)$)  each step. A Neural Statistician instead feeds a full set of elements $X$ through both encoder ($q(c;X)$) and decoder ($p(X|c)$) networks, in order to share a latent variable $c$. In a VHE, we bound the full likelihood $p(X)$ using only random subsamples $D$ and $x$ for encoding/decoding. Optionally, the decoder $p(x|c)$ may be defined through a local latent variable $z$.}
\end{figure*}

Much research has focused on developing neural architectures for few-shot classification \citep{koch2015siamese,vinyals2016matching,snell2017prototypical,santoro2016one}. These discriminatively-trained networks take as input a test example and a `support set' of examples from several novel classes, and determine the most likely classification of the test example within the novel classes. A second approach, as explored in \citet{ravi2016optimization,finn2017model}, is to use only a standard classification network but adapt its parameters to the support examples with a learned initialisation and update rule. In either case, such discriminative models can achieve state-of-the-art few-shot classification performance, although they provide no principled means for transferring knowledge to other tasks.

An alternative approach centers on few-shot learning of \textit{generative} models, from which good classification ought to come for free. Much recent work on meta-learning aims to take one or a few observations from a set $D$ as input, and produce a distribution over new elements $p(x|D)$ by some learning procedure, expressed either as a neural network 
\citep{rezende2016one,bartunov2016fast,reed2017few}
or by adapting the parameters of an unconditional model \citep{reed2017few}.

A promising route to learning generative models is hierarchical Bayesian inference, which aims to capture shared structure between instances through \emph{shared} latent variables. A recent example is developed in \citet{lake2015human}: a compositional, causal generative model of handwritten characters which achieves state-of-the-art results at few-shot character classification, alphabet classification, and both conditional and unconditional generation.
However, this model was hand engineered for the Omniglot domain, and so leaves open the challenge of how to learn such hierarchical Bayesian models using only a generic architecture.
The recently proposed \textit{Neural Statistician} \citep{edwards2016towards} offers one means towards this, using amortised variational inference to support learning in a deep hierarchical generative model.

In this work we aim to learn generative models, expressed using high capacity neural network architectures, from just a few examples of a concept. To this end we propose the \textit{Variational Homoencoder} (VHE), combining several advantages of the models described above:

\begin{enumerate}
\item Like conditional generative approaches (\textit{e.g.} \citet{rezende2016one}), we train on a few-shot generation objective which matches how our model may be used at test time. However, by introducing an encoding cost, we simultaneously optimise a likelihood lower bound for a hierarchical generative model, in which structure shared across elements is made explicit by shared latent variables.

\item \citet{edwards2016towards} has learned hierarchical Bayesian models by applying Variational Autoencoders to sets, such as classes of images. However, their approach requires feeding a full set through the model per gradient step (Figure 1), rendering it intractable to train on very large sets. In practice, they avoid computational limits by training on smaller, random subsets. In a VHE, we instead optimise a likelihood bound for the complete dataset, while \textit{constructing this bound} by subsampling. This approach can not only improve generalisation, but also departs from previous work by extending to models with richer latent structure, for which the joint likelihood cannot be factorised.

\item As with a VAE, the VHE objective includes both an \textit{encoding-} and \textit{reconstruction-} cost. However, by sharing latent variables across a large set of elements, the \textit{encoding cost} per element is reduced significantly. This facilitates use of powerful autoregressive decoders, which otherwise often suffer from ignoring latent variables \citep{chen2016variational}. We demonstrate the significance of this by applying a VHE to the Omniglot dataset. Using a PixelCNN decoder \citep{oord2016conditional}, our generative model is arguably the first with a general purpose architecture to both attain near state-of-the-art one-shot classification performance and produce high quality samples in one-shot generation.
\end{enumerate}

\section{BACKGROUND}

\subsection{VARIATIONAL AUTOENCODERS}
\begin{algorithm*}[t]
\label{alg:vhetraining}
  \hspace{1em} initialize $(\theta, \phi)$ \hfill \textit{Parameters for decoder $p$ and encoder $q$}\\
  \hspace{1em} \textbf{repeat}\\
	\hspace{2em} sample $(x_k, i_k)$ for $k=1,\ldots,M$ \hfill \textit{Minibatch of elements with corresponding class labels}\\
    \hspace{2em} sample $D_k \subseteq X_{i_k}$ for $k=1,\ldots,M$\hfill \textit{where }$|D_k|=N$\\
    \hspace{2em} sample $c_k \sim q_\phi(c;D_k)$ for $k=1,\ldots,M$\\
    \hspace{2em} \textit{(optional)} sample $z_k \sim q_\phi(z;c_k,x_k)$ for $k=1,\ldots,M$\\
    \hspace{2em} $\mathbf{g} \approx \frac{1}{M} \sum_k \nabla \mathcal{L}_{\theta, \phi}(x_k;D_k,|X_{i_k}|)$ \hfill \textit{Reparametrization gradient estimate using $\mathbf{c}, \mathbf{z}$}\\
    \hspace{2em} $(\theta, \phi) \leftarrow (\theta, \phi) + \lambda\mathbf{g}$ \hfill \textit{Gradient step, e.g SGD} \\    
  \hspace{1em} \textbf{until} convergence of $(\theta, \phi)$
  \caption{Minibatch training for the \textit{Variational Homoencoder}.
  Minibatches are of size $M$. Stochastic inference network $q$ uses subsets of size $N$.}
\end{algorithm*}
When dealing with latent variable models of the form $p(x) = \int_z p(z)p(x|z)\textnormal{d}z$,
the integration is necessary for both learning and inference but is often intractable to compute in closed form. \textit{Variational Autoencoders} (VAEs, \citet{kingma2013auto}) provide a method for learning such models by utilising neural-network based approximate posterior inference. Specifically, a VAE comprises a generative network $p_\theta(z)p_\theta(x|z)$ alongside a separate inference network $q_\phi(z;x)$. These are trained jointly to maximise a single objective:

\begin{align}
&\mathcal{L}_X(\theta, \phi) = \nonumber \\
&\sum_{x \in X} \left[ \textnormal{log }p_\theta(x) - \textnormal{D}_{KL}\Big( q_\phi(z;x) \parallel p_\theta(z|x) \Big) \right] \label{eq:vae1}\\
&= \sum_{x \in X} \left[ \mathop{\mathbb{E}}_{q_\phi(z; x)} \textnormal{log } p_\theta(x|z) - \textnormal{D}_{KL}\Big( q_\phi(z;x) \parallel p_\theta(z) \Big) \right] \label{eq:vae2}
\end{align}

As can be seen from Equation \ref{eq:vae1}, this objective $\mathcal{L}_X$ is a lower bound on the total log likelihood of the dataset $\sum_{x \in X}\textnormal{log }p_\theta(x)$, while $q_\phi(z;x)$ is trained to approximate the true posterior $p_\theta(z|x)$ as accurately as possible. If it could match this distribution exactly then the bound would be tight so that the VAE objective equals the true log likelihood of the data. In practice, the resulting model is typically a compromise between two goals: pulling $p_\theta$ towards a distribution that assigns high likelihood to the data, but also towards one which allows accurate inference by $q_\phi$. Equation \ref{eq:vae2} provides a formulation for the same objective which can be optimised stochastically, using Monte-Carlo integration to approximate the expectation. For brevity, we will omit subscripts $\theta,\phi$ for the remainder of this paper.

\subsection{VARIATIONAL AUTOENCODERS OVER SETS}

The \textit{Neural Statistician} \citep{edwards2016towards} is a Variational Autoencoder in which each item to be encoded is itself a set, such as the set $X^{(i)}$ of all images with a particular class label $i$:
\begin{align}
X^{(i)} = \{x^{(i)}_1, x^{(i)}_2, \cdots, x^{(i)}_n\}
\end{align}
The generative model for sets, $p(X)$, is described by introduction of a corresponding latent variable $c$. Given $c$, individual $x \in X$ are conditionally independent:
\begin{align}
p(X) = \int_{c} p(c)  \prod_{x \in X} p(x | c) \mathrm{d}c
\end{align}
The likelihood is again intractable to compute, but it can be bounded below via:
\begin{align}
\label{lower-bound-neural-statistician}
&\log p(X) \ge \mathcal{L}_X = \nonumber \\
&\mathop{\mathbb{E}}_{q(c;X)} \bigg[ \sum_{x \in X} \log p(x | c ) \bigg] - \textnormal{D}_{KL}\big(q(c;X) \parallel p(c)\big)
\end{align}

Unfortunately, calculating the variational lower bound for each set $X$ requires evaluating both $q(c; X)$ and $p(X | c)$, meaning that the entire set must be passed through both networks for each gradient update. This can become computationally challenging for classes with hundreds of examples. Instead, previous work \citep{edwards2016towards} ensures that sets used for training are always of small size by maximising a log-likelihood bound for randomly sampled subsets $D \subset X$:
\begin{align}
\label{lower-bound-neural-statistician-D}
&\mathop{\mathbb{E}}_{D \subset X} \Bigg[ \mathop{\mathbb{E}}_{q(c;D)} \bigg[ \sum_{x \in D} \log p(x | c ) \bigg] - \textnormal{D}_{KL}\big(q(c;D) \parallel p(c)\big) \Bigg]
\end{align}

As we demonstrate in section \ref{expts}, this subsampling decreases the model's incentive to capture correlations within a class, reducing utilisation of the latent variables. This poses a significant challenge when scaling up to more powerful generative networks, which require a greater incentive to avoid simply memorising the global distribution. Our work addresses this by replacing the variational lower-bound in Equation \ref{lower-bound-neural-statistician-D} with a new objective, which better incentivises the use of latent variables, leading to improved generalisation.

\section{VARIATIONAL HOMOENCODERS}


Rather than bound the likelihood of subsamples $D$ from a set, as in \citet{edwards2016towards}, we instead use subsampling to construct a lower bound on the complete set $X$. We use a constrained variational distribution $q(c; D), D \subseteq X$ for posterior inference and an unbiased stochastic approximation $\text{log }p(x | c), x \in X$ for the likelihood. This bound will typically be loose due to stochasticity in sampling $D$, and we view this as a regularization strategy: we aim to learn latent representations that are quickly inferable from a small number of instances, and the VHE objective is tailored for this purpose. 


\subsection{STOCHASTIC LOWER BOUND}\label{lower-bound}

\begin{figure*}[t]
\label{fig:structure}
\center
\includegraphics[width=0.9\textwidth]{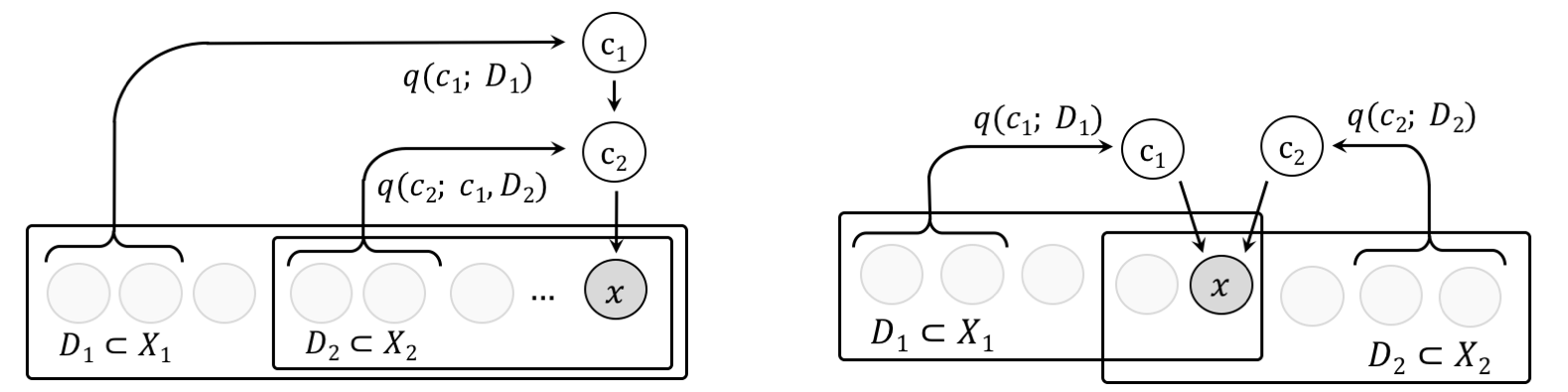}
\caption{Application of VHE framework to hierarchical (\textit{left}) and factorial (\textit{right}) models. Given an element $x$ such that $x \in X_1$ and $x \in X_2$, an approximate posterior is constructed for the corresponding shared latent variables $c_1, c_2$ using subsampled sets $D_1 \subset X_1, D_2 \subset X_2$.}
\end{figure*}

We would like to learn a generative model for sets $X$ of the form
\begin{align}
p(X) = \int p(c)\prod_{x \in X}p(x|c) \textnormal{d}c
\end{align}

We will refer our full dataset as a union of disjoint sets $\mathcal{X} = X_1 \sqcup X_2 \sqcup \ldots \sqcup X_n$, and use $X_{(x)}$ to refer to the set $X_i \ni x$. Using the standard consequent of Jensen's inequality, we can lower bound the log-likelihood of each set $X$ using an arbitrary distribution $q$. In particular, we give $q$ as a fixed function of arbitrary data.
\begin{align}
\textnormal{log }p(X) \ge \mathop{\mathbb{E}}_{q(c;D)} \textnormal{log }p(X|c)  - \textnormal{D}_{KL}\big[q(c;D) \parallel p(c)\big], & \nonumber
\\ \quad \forall D \subset X& 
\end{align}

Splitting up individual likelihoods, we may rewrite
\begin{align}
\textnormal{log }p(X) &\ge \mathop{\mathbb{E}}_{q(c;D)}\Big[\sum_{x \in X}\textnormal{log }p(x|c)\Big] \nonumber \\
&- \textnormal{D}_{KL}\big[q(c;D) \parallel p(c)\big], \quad &\forall D \subset X\\
\label{eq:klrescaling}
&= \sum_{x \in X}\Big[ \mathop{\mathbb{E}}_{q(c;D)} \textnormal{log } p(x|c) \nonumber \\
&- \frac{1}{|X|} \textnormal{D}_{KL}\big[q(c;D) \parallel p(c)\big] \Big], \quad &\forall D \subset X\\
&\mathop{=}^\textrm{def} \sum_{x \in X} \mathcal{L}(x;D,|X|), \quad &\forall D \subset X \label{eq:L}
\end{align}

Finally, we can replace the universal quantification with an expectation under any distribution of D (e.g. uniform sampling from X without replacement):
\begin{align}
\label{eq:vheobjectiveX}
\textnormal{log }p(X) &\ge \mathop{\mathbb{E}}_{D \subset X} \sum_{x\in X} \mathcal{L}(x;D,|X|) \\
&= \sum_{x\in X} \mathop{\mathbb{E}}_{D \subset X} \mathcal{L}(x;D,|X|) \\ \label{eq:vheobjective}
\textnormal{log }p(\mathcal{X}) &\ge \sum_{x\in \mathcal{X}} \mathop{\mathbb{E}}_{D \subset X_{(x)}} \mathcal{L}(x;D,|X_{(x)}|)
\end{align}
This formulation suggests a simple modification to the VAE training procedure, as shown in Algorithm 1. At each iteration we select an element $x$, use resampled elements $D \subset X_{(x)}$ to construct the approximate posterior $q(c;D)$, and rescale the encoding cost appropriately.
\begin{align}
\label{eq:vheobjective_new}
&\textit{VHE objective:} \nonumber \\
&\mathop{\mathbb{E}}_{\subalign{
x &\in \mathcal{X}\\ D &\subset X_{(x)}}} \Bigg[ \mathop{\mathbb{E}}_{q(c;D)}  \textnormal{log }p(x|c) - \frac{1}{\vert X_{(x)} \vert} \textnormal{D}_{KL}\big[q(c;D) \parallel p(c)\big] \Bigg]
\end{align}
If the generative model $p(x|c)$ also describes a separate latent variable $z$ for each element, we may simply introduce a second inference network $q(z;c,x)$ in order to further bound the reconstruction error of Equation \ref{eq:vheobjective_new}:

\textit{VHE objective with per-element latent variables:} 
\begin{align}
\label{eq:vheobjective_z}
\mathop{\mathbb{E}}_{\subalign{
x &\in \mathcal{X}\\ D &\subset X_{(x)}}} \Bigg[ &\mathop{\mathbb{E}}_{\substack{q(c;D)\\ q(z;c,x)}} \textnormal{log } p(x|c,z) - \textnormal{D}_{KL}\big[ q(z;c,x) \parallel p(z|c) \big] \nonumber \\
&- \frac{1}{\vert X_{(x)} \vert} \textnormal{D}_{KL}\big[q(c;D) \parallel p(c)\big] \Bigg]
\end{align}


\subsection{APPLICATION TO STRUCTURED DATASETS}

The above derivation applies to a dataset partitioned into \textit{disjoint} subsets $\mathcal{X} = X_1 \sqcup X_2 \sqcup \ldots \sqcup X_n$, each with a corresponding latent variable $c_i$. However, many datasets offer a richer organisational structure, such as the hierarchical grouping of characters into alphabets \citep{lake2015human} or the factorial categorisation of rendered faces by identity, pose and lighting \citep{kulkarni2015deep}.

Provided that such organisational structure is known in advance, we may generalise the training objective in Equation \ref{eq:vheobjective} to include a separate latent variable $c_i$ for each group $X_i$ within the dataset, even when these groups overlap. To do this we first rewrite this bound in its most general form, where $\mathbf{c}$ collects all latent variables:
\begin{align}
\textnormal{log }p(\mathcal{X}) &\ge \mathop{\mathbb{E}}_{Q(\mathbf{c};\mathbf{D})}\Big[\sum_{x \in X}\textnormal{log }p(x|\mathbf{c})\Big] \nonumber \\
& - \textnormal{D}_{KL}\big[Q(\mathbf{c};\mathbf{D}) \parallel P(\mathbf{c})\big]
\end{align}
As shown in Figure 2, a separate $D_i \subset X_i$ may be subsampled for inference of each latent variable $c_i$, so that $Q(\mathbf{c}) = \prod_i q_i(c_i;D_i)$. This leads to an analogous training objective (Equation \ref{eq:structure}), which may be applied to data with factorial or hierarchical category structure. For the hierarchical case, this objective may be further modified to infer layers sequentially, derived in Supplementary Material.
\begin{align}
\label{eq:structure}
\textnormal{log }p(\mathcal{X}) &\ge \sum_{x \in \mathcal{X}}\mathop{\mathbb{E}}_{\substack{D_i \subset X_i \\ \textrm{for each} \\ i: x \in X_i}} \Bigg[ \mathop{\mathbb{E}}_{\substack{q_i(c_i;D_i) \\ \textrm{for each} \\ i:x \in X_i}} \textnormal{log }p(x|\mathbf{c}) \nonumber \\
&- \sum_{i:x \in X_i} \frac{1}{\vert X_i \vert} \textnormal{D}_{KL}\big(q_i(c_i;D_i) \parallel p(c_i)\big)\Bigg]
\end{align}

\subsection{POWERFUL DECODER MODELS}
As evident in Equation \ref{eq:klrescaling}, the VHE objective provides a formal motivation for KL rescaling in the variational objective (a common technique to increase use of latent variables in VAEs) by sharing these variables across many elements. This is of particular importance when using autoregressive decoder models, for which a common failure mode is to learn a decoder $p(x|z)$ with no dependence on the latent space, thus avoiding the encoding cost. In the context of VAEs, this particular issue has been discussed by \citet{chen2016variational} who suggest crippling the decoder as a potential remedy. 

The same failure mode can occur when training a VAE for sets,
particularly when the sets $D$ are of small size and thus have low total correlation.
\textit{Variational Homoencoders} suggest a potential remedy to this, encouraging use of the latent space by reusing the same latent variables across a large set $X$. This allows a VHE to learn useful representations even with $|D|=1$, while at the same time utilising a powerful decoder to achieve highly accurate density estimation. In our experiments, we exploit the VHE's ability to use powerful decoders: specifically, we learn a generative model with a PixelCNN decoder, which is not possible with previous frameworks.

\begin{figure}[t]
\center\includegraphics[width=0.35\textwidth]{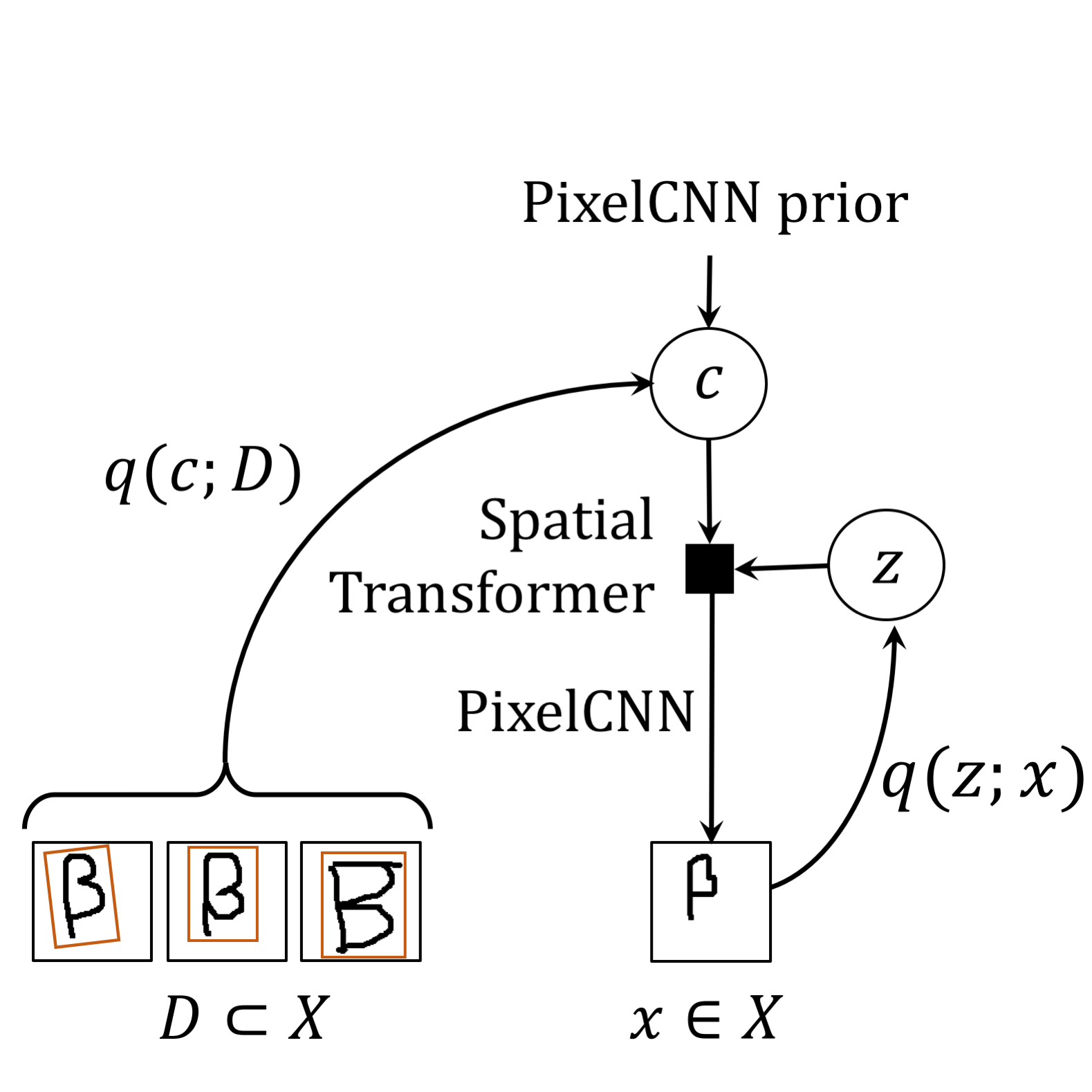}
\label{fig:architecture}
\caption{PixelCNN \textit{VHE} architecture used for Omniglot and Youtube Faces. A spatial transformer network $q(c;D)$ encodes a subset $D$ of a character class into a class latent variable $c$ with the same width and height as the input image. A separate encoder $q(z;x)$, parametrized by a convolutional network, encodes position information in the target image into a latent variable $z$. A PixelCNN prior is used for $c$, and a Gaussian prior for $z$. During decoding, $c$ and $z$ are combined by a spatial transformer and used to condition a PixelCNN decoder network $p(x|\textrm{STN}(c,z))$.}
\end{figure}

\section{EXPERIMENTAL RESULTS}
\label{expts}

\subsection{HANDWRITTEN CHARACTER CLASSES}
\label{expt1}

To demonstrate that the VHE objective can facilitate learning with more expressive generative networks, we trained a variety of models on the Omniglot dataset exploring the interaction between model architecture and training objective. We consider two model architectures: a standard deconvolutional network based on \citet{edwards2016towards}, and a hierarchical PixelCNN architecture inspired by the PixelVAE \citep{gulrajani2016pixelvae}. For each, we compare models trained with the VHE objective against three alternative objectives.

For our hierarchical PixelCNN architecture (Figure 3) each character class is associated with a spatial latent variable $c$ (a character `template') with a PixelCNN prior, and each image $x$ is associated with its own latent variable $z$ (its `position') with a Gaussian prior. To generate $x$, a Spatial Transformer Network (STN) \citep{jaderberg2015spatial} applies $z$ to the class template $c$, and the result is input to a Gated PixelCNN $p(x|\textrm{STN}(c,z))$ \citep{oord2016conditional}. The position encoder $q(z; x)$ is given by a CNN, and the class encoder $q(c;D)$ by an STN averaged over $D$. Both produce diagonal Gaussian distributions.

Using both PixelCNN and deconvolutional architectures, we trained models by several objectives. We compare a VHE model against a Neural Statistician baseline, with each trained on sampled subsets $D \subset X$ with $|D|=5$ (as in \citet{edwards2016towards}). Secondly, since the VHE introduces both data-resampling and KL-rescaling as modifications to this baseline, we separate the contributions of each using two intermediate objectives:

\begin{table}[t!]
\centering 
  \caption{Comparison of VHE, Neural Statistician, and intermediate objectives with both deconvolutional and PixelCNN architectures. Rescaling encourages use of the latent space, while resampling encourages generalisation from the support set. The VHE is able to utilise the PixelCNN to achieve the highest classification accuracy.}
  \begin{tabular}{lrrr}
    \toprule
    \cmidrule{1-2}
         \multicolumn{2}{r}{KL / nats*} & \multicolumn{1}{l}{Accuracy}\\
         & & (5-shot)\\
    \midrule
\multicolumn{3}{l}{\textbf{Deconvolutional Architecture} }\\
    \rule{0pt}{0.5cm}Neural Statistician [\citenum{edwards2016towards}] & 31.34 & 95.6\%\\
    Resample (Eq \ref{rsm})& 25.74 & 94.0\%\\
    Rescale (Eq \ref{rsc})& 477.65 & 95.3\%\\
    \textit{VHE} (resample + rescale, Eq \ref{eq:vheobjective_z}) & 452.47 & 95.6\%    \\
    \midrule
\multicolumn{3}{l}{\textbf{PixelCNN Architecture}}\\
    \rule{0pt}{0.5cm}Neural Statistician & 14.90 & 66.0\%\\
    Resample & 0.22 & 4.9\%\\
    Rescale & 506.48 & 62.8\%\\
    \textit{VHE} (resample + rescale) & 268.37 & \textbf{98.8\%}    \\
    \bottomrule
    \multicolumn{3}{r}{\rule{0pt}{0.4cm}*$\textnormal{D}_{KL}\big(q(c;D) \parallel p(c)\big)$, train set}
  \end{tabular}
\end{table}

\begin{align}
&\textit{\footnotesize{Resample only:}} \nonumber\\
& \underbrace{\mathop{\mathbb{E}}_{\substack{
D \subset X\\ x \in X}}}_{\substack{\textrm{resample}\\ \textrm{decoded}\\ \textrm{element}}} \Bigg[ \mathop{\mathbb{E}}_{q(c;D)}  \textnormal{log }p(x|c)- \frac{1}{\vert D \vert} \textnormal{D}_{KL}\big[q(c;D) \parallel p(c)\big] \Bigg] \label{rsm} \\
&\textit{\footnotesize{Rescale only: \nonumber}} \\
&\mathop{\mathbb{E}}_{\substack{
D \subset X\\ x \in D}} \Bigg[ \mathop{\mathbb{E}}_{q(c;D)}  \textnormal{log }p(x|c)-\underbrace{\frac{1}{\vert X \vert}}_{\text{rescale KL}} \textnormal{D}_{KL}\big[q(c;D) \parallel p(c)\big] \Bigg] \label{rsc}
\end{align} 

All models were trained on a random sample of 1200 Omniglot classes using images scaled to 28x28 pixels, dynamically binarised, and augmented by 8 rotations/reflections to produce new classes. We additionally used 20 small random affine transformations to create new instances within each class. Models were optimised using Adam \citep{kingma2013auto}, and we used training error to select the best parameters from 5 independent training runs. We also implemented the `sample dropout' trick of \citet{edwards2016towards}, but found that this had no effect on performance. At test time we classify an example $x$ by Monte Carlo estimation of the expected conditional likelihood under the variational posterior $E_{q(c;D)}p(x|c)$, with 20 samples from $q(c;D)$. $x$ is then classified to class with support set D that maximises this expected conditional likelihood.

\begin{table}[t!]
  \caption{Comparison of classification accuracy with previous work. The VHE objective allows us to use a powerful decoder network, yielding state-of-the-art few-shot classification amongst deep generative models.}
  \label{table:classification}
  \begin{tabular}{lrr}
    \toprule
    \cmidrule{1-3}
       \multicolumn{3}{r}{Classification Accuracy (20-way)} \\
         &\multicolumn{1}{c}{1-shot} & \multicolumn{1}{c}{5-shot}\\
    \midrule
    \multicolumn{3}{l}{\textbf{Generative models}, $\textnormal{log }p(X)$}\\
    Generative Matching Networks [\citenum{bartunov2016fast}] & 77.0\% & 91.0\%\\
    Neural Statistician [\citenum{edwards2016towards}] & 93.2\% & 98.1\%\\
    \textit{VHE} & \textbf{95.2\%} & \textbf{98.8\%}\\
    \midrule
    \multicolumn{3}{l}{\textbf{Discriminative models}, $\textnormal{log }q(y;x,X,Y)$}\\
    Matching Networks [\citenum{vinyals2016matching}] & 93.8\% & 98.7\% \\
    Convnet with memory module [\citenum{kaiser2017learning}]& 95.0\% & 98.6\% \\ 
    mAP-DLM [\citenum{triantafillou2017few}] & 95.4\% & 98.6\% \\
    Model-Agnostic Meta-learning [\citenum{finn2017model}] & 95.8\% & \textbf{98.9}\% \\  
    Prototypical Networks [\citenum{snell2017prototypical}] & \textbf{96.0}\% & \textbf{98.9}\% \\     
    \midrule
\textit{(VHE, within-alphabet\footnotemark)} & 81.3\% & 90.3\%\\
    \end{tabular}%
\end{table}

\footnotetext{The few-shot classification task defined by \citet{lake2015human} is to identify an image to one of 20 character classes, where \textit{all 20 classes belong to the same (unseen) alphabet}. However, most work since has evaluated on an easier one-shot classification task, in which the 20 support characters are drawn from the entire test set (so are typically more dissimilar). We find that our model performs significantly worse on the within-alphabet variant, and so include results to facilitate future comparison on this more challenging task. Attaining near-human classification accuracy on this variant remains an open challenge for neural network models.}

\begin{figure*}[h]
\center\colorbox{white}{\includegraphics[width=\textwidth]{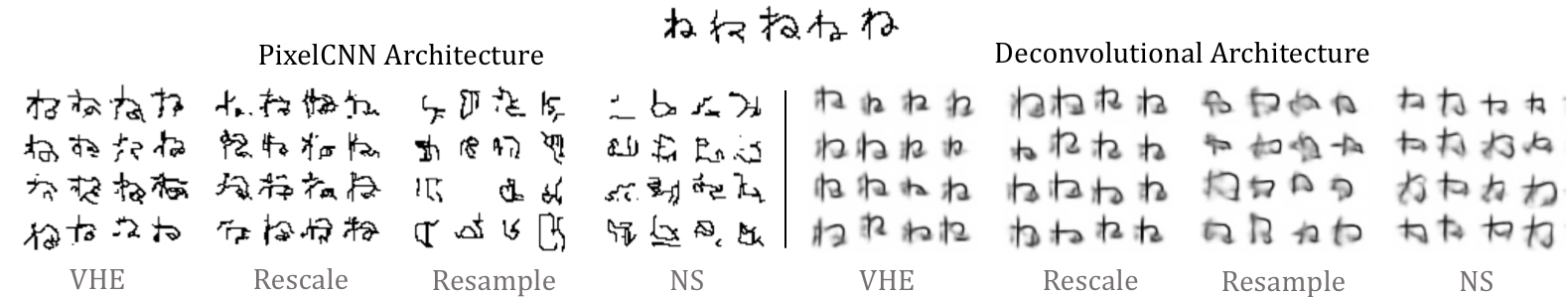}}
\label{fig:octants}
\caption{5-shot samples generated by each model (more in Supplement). With a PixelCNN architecture, both Neural Statistician and Resample objectives lead to underutilisation of the latent space, producing unfaithful samples.}
\end{figure*}

Table 1 collects classification results of models trained using each of the four alternative training objectives, for both architectures. For a deconvolutional architecture, we find little difference in classification performance between all four training objectives, with the Neural Statistician and VHE models achieving equally high accuracy.

For the hierarchical PixelCNN architecture, however, significant differences arise between training objectives. In this case, a Neural Statistician learns a strong global distribution over images but makes only minimal use of latent variables $c$. This means that, despite the use of a higher capacity model, classification accuracy is much poorer (66\%) than that achieved using a deconvolutional architecture. For the same reason, conditional samples display an improved sharpness but are no longer identifiable to the cue images on which they were conditioned (Figure 4). Our careful training suggests that this is not an optimisation difficulty but is core to the objective, as discussed in \citet{chen2016variational}.
 
By contrast, a VHE is able to gain a large benefit from the hierarchical PixelCNN architecture, with a 3-fold reduction in classification error (5-shot accuracy 98.8\%) and conditional samples which are simultaneously sharp and identifiable (Figure 4). This improvement is in part achieved by increased utilisation of the latent space, due to rescaling of the KL divergence term in the objective. However, our results show that this common technique is insufficient when used alone, leading to overfitting to cue images with an equally severe impairment of classification performance (accuracy 62.8\%). Rather, we find that KL-rescaling and data resampling must be used together in order for the benefit of the powerful PixelCNN architecture to be realised.

\begin{figure}[h]
\label{fig:oneshot}
\center
\includegraphics[trim={0 22cm 0 0},clip,width=0.42\textwidth]{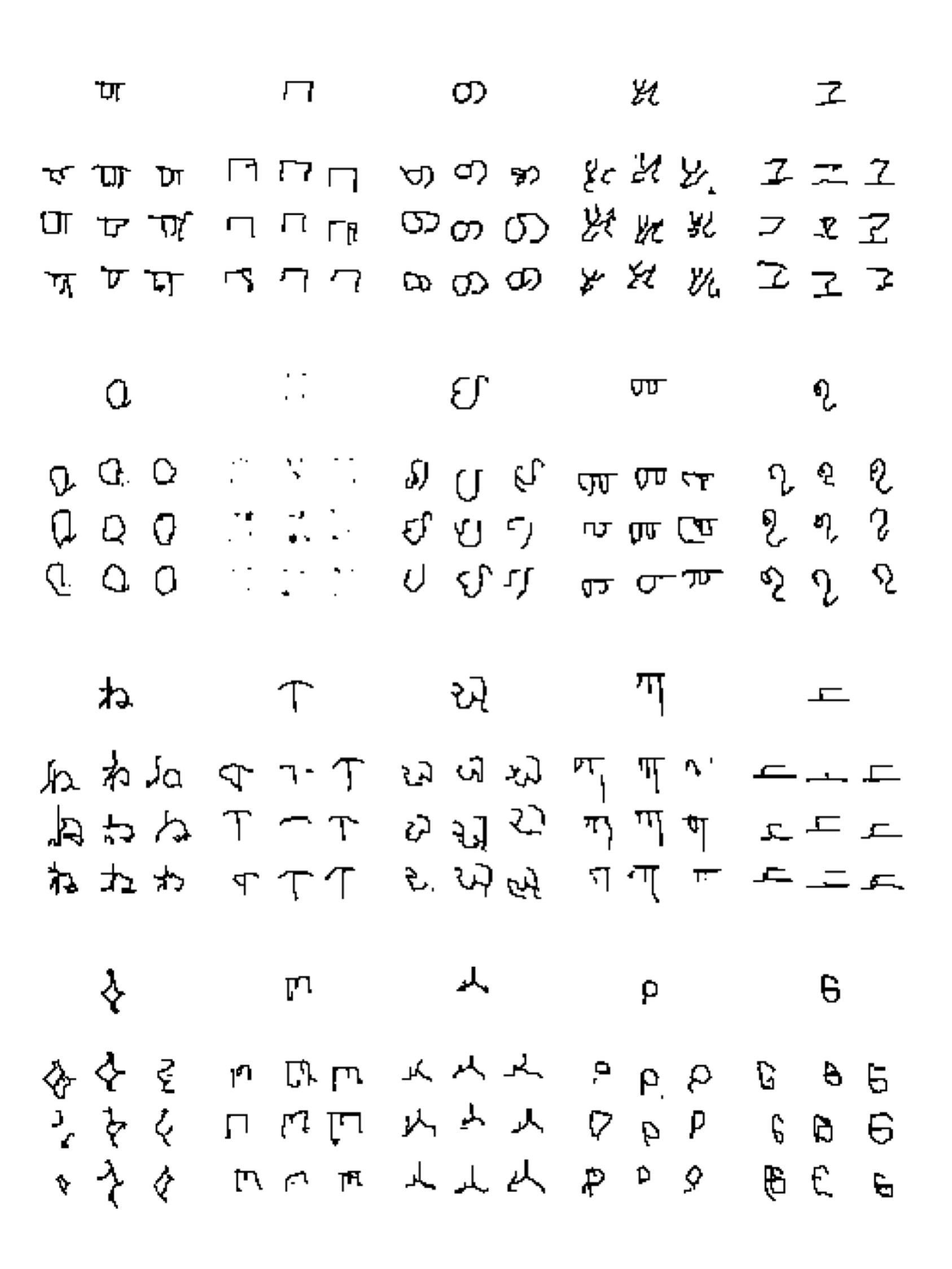}
\caption{One-shot same-class samples generated by our model. Cue images were sampled from previously unseen classes.}
\end{figure}

Table 2 lists the classification accuracy achieved by VHEs with both $|D|=1$ and $|D|=5$, as compared to existing deep learning approaches. We find that both networks are not only state-of-the-art amongst deep generative models, but also competitive against the best discriminative models trained directly for few-shot classification. Unlike these discriminative models, a VHE is also able to generate new images of a character in one shot, producing samples which are both realistic and faithful to the class of the cue image (Figure 5).

As our goal is to model shared structure across images, we evaluate generative performance using joint log likelihood of the entire Omniglot test set (rather than separately across images). From this perspective, a single element VAE will perform poorly as it treats all datapoints as independent, optimising a sum over log likelihoods for each element. By sharing latents across elements of the same class, a VHE can improve upon this considerably. 

For likelihood evaluation, our most appropriate comparison is with Generative Matching Networks \citep{bartunov2016fast} as they also model dependencies within a class. Thus, we trained models under the same train/test split as them, with no data augmentation. We evaluate the joint log likelihood of full character classes from the test set, normalised by the number of elements, using importance weighting with k=500 samples from $q(c;X)$. As can be seen in Tables 3 and 4, our hierarchical PixelCNN architecture is able to achieve state-of-the-art log likelihood results only when trained using the VHE objective.

\begin{table*}[t]
\begin{minipage}[t]{.45\textwidth}
\centering
	    \captionof{table}{Joint NLL of Omniglot test set, compared across architectures and objectives.}
  \begin{tabular}{lrr}
    \toprule
    \cmidrule{1-2}
         & Test NLL per image\\
    \midrule
\multicolumn{2}{l}{\textbf{Deconvolutional Architecture}}\\
    \rule{0pt}{0.5cm}NS [\citenum{edwards2016towards}]  & 102.84 nats\\
    Resample & 110.30 nats\\
    Rescale  & 109.01 nats\\
    \textit{VHE (resample + rescale)} & 104.67 nats\\
    \midrule
\multicolumn{2}{l}{\textbf{PixelCNN Architecture}}\\
    \rule{0pt}{0.5cm}NS  & 73.50 nats\\
    Resample  & 66.42 nats\\
    Rescale  & 71.37 nats \\
    \textit{VHE (resample + rescale)} & \textbf{61.22 nats}    \\
    \bottomrule
  \end{tabular}
\end{minipage}
\hspace{.03\textwidth}
\begin{minipage}[t]{0.54\textwidth}
\centering
\label{table:likelihood}
    \captionof{table}{Comparison of deep generative models by joint NLL of Omniglot test set.}
  \begin{tabular}{lr}
    \toprule
    	         \multicolumn{2}{r}{Test NLL per image}\\
        \midrule
    \multicolumn{1}{l}{\textbf{Independent models}} & $\frac{1}{n}\textnormal{log }\prod_i p(x_i)$\\
\rule{0pt}{0.5cm}DRAW [\citenum{gregor2015draw}] & $< 96.5$ nats\\
    Conv DRAW [\citenum{gregor2016towards}]& $< 91.0$ nats \\
    VLAE [\citenum{chen2016variational}] & 89.83 nats\\
   \cmidrule{1-2}
    \multicolumn{2}{l}{\textbf{Conditional models} \hfill $\frac{1}{n}\textnormal{log }\prod_i p(x_i | x_{1:i-1})$}  \\
\multicolumn{2}{l}{\rule{0pt}{0.5cm}Generative Matching Networks [\citenum{bartunov2016fast}] \hfill 62.42 nats\footnotemark}\\
   \cmidrule{1-2}
    \multicolumn{1}{l}{\textbf{Shared-latent models}} & $\frac{1}{n}\textnormal{log }\mathbb{E}_{p(c)}\prod_i p(x_i|c)$\\
    \rule{0pt}{0.5cm}\textit{Variational Homoencoder} & \textbf{61.22 nats}\\
    \bottomrule\\
  \end{tabular}
\end{minipage}
\end{table*}

\subsection{YOUTUBE FACES}

To confirm that our approach can be used to produce naturalistic images, we compare VHE and Neural Statistician models trained on images from the YouTube Faces Database\citep{wolf2011face}, comprising 3,425 videos of 1,595 celebrities downloaded from YouTube. For our experiments, we use the aligned and cropped to face version, additionally cropping each image by 50\% in both height and width, and rescale to 40x40 pixels. Our training, validation, and test sets consist of one video per person and 48 images per video. We use 954 videos for the training set and 641 videos for the test set. 

We consider two architectures: the hierarchical PixelCNN network used for Omniglot experiments, and the deconvolution network used to model faces in \citet{edwards2016towards}. As above, we train each model using both VHE and NS objectives with $|D| = 5$. 

\footnotetext{We thank the authors of \citet{bartunov2016fast} for providing us with this comparison.}

\begin{table}[h]
  \caption{Classification results for YouTube faces dataset. The VHE PixelCNN utilises the latent space most effectively, and therefore achieves the highest few-shot classification accuracy and test image NLL.}
    \begin{tabular}{lrrr}
    \toprule
    \cmidrule{1-3}
    & Test NLL & \multicolumn{2}{r}{Accuracy (200-way)} \\
      & per image &\multicolumn{1}{c}{1-shot} & \multicolumn{1}{c}{5-shot}\\
    \midrule
    \multicolumn{3}{l}{\textbf{Deconvolutional}}\\
    Neural Statistician & 12512.4 & 39.2\% & 49.0\%\\
    \textit{VHE} & 12717.6 & 37.2\% & 44.8\%\\
    \midrule
    {\textbf{PixelCNN}}\\
    Neural Statistician & 4229.8 & 92.1\% & 98.5\%\\
    \textit{VHE} & \textbf{4091.3} &\textbf{92.5\%} & \textbf{98.9\%}\\
    \bottomrule
    \end{tabular}%
\label{table:faces_table}
\end{table}  

\begin{figure*}[h]
\centering
\includegraphics[trim={0 1cm 0 0},clip,width=\textwidth]{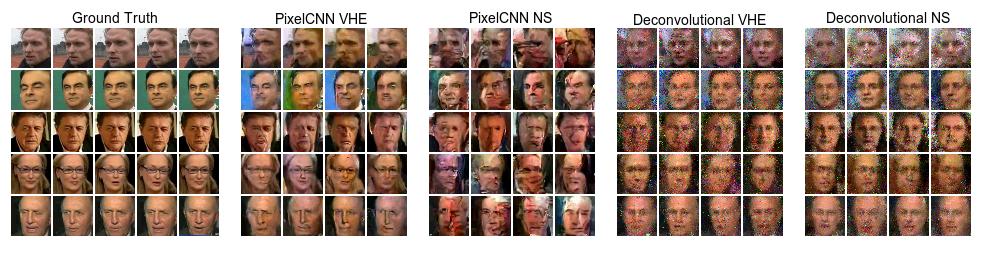}
\caption{5-shot samples of YouTube faces generated using both PixelCNN and deconvolutional architectures. Note that, for accurate comparison, we \textit{sample} images from the decoder rather than taking the conditional mode as is common. For the deconvolutional models, this leads to images which appear more noisy than shown in previous work.}
\label{fig:facesimages}
\end{figure*}

Classification results for trained models are shown in Table \ref{table:faces_table}, and conditionally generated samples in Figure \ref{fig:facesimages}. As with Omniglot experiments, we find that the VHE objective improves use of the hidden layer $c$, leading to more accurate classification and conditional generation than the Neural Statistician. While the deconvolutional architecture is capable of producing realistic images (see \citet{edwards2016towards}), our results show that it is not powerful enough to perform accurate few-shot classification. On the other hand, the PixelCNN architecture trained using the Neural Statistician objective achieves accurate few-shot classification, but generates poor images. The only network able to produce realistic images \emph{and} perform accurate classification is the PixelCNN trained using our VHE objective.

\subsection{MODELLING RICH CATEGORY STRUCTURE}
\begin{figure}[t]
\includegraphics[width=0.45\textwidth]{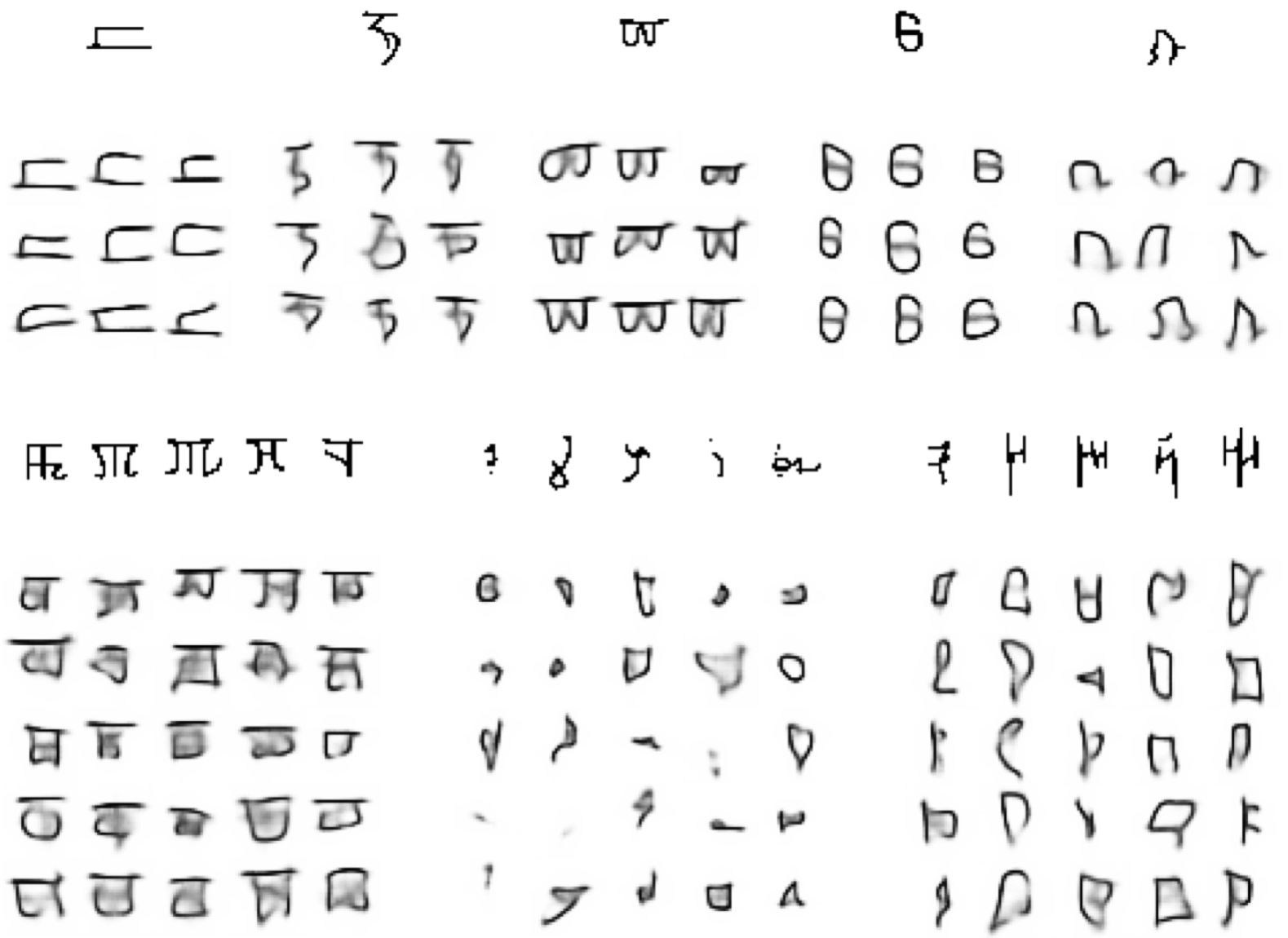}
\caption{ Conditional samples from character (top) and alphabet (bottom) levels of the same hierarchical model.}
\label{fig:alphabets}
\end{figure}

To demonstrate how the VHE framework may apply to models with richer category structure, we built both a hierarchical and a factorial VHE (Figure 2) using simple modifications to the above architectures. For the hierarchical VHE, we extended the deconvolutional model with an extra latent layer $a$ using the same encoder and decoder architecture as $c$. This was used to encode alphabet level structure for the Omniglot dataset, learning a generative model for alphabets of the form
\begin{align}
p(\mathcal{A}) = \int p(a) \prod_{X_i \in \mathcal{A}} \int p(c_i | a)\prod_{x_{ij} \in X_i} p(x_{ij}|c_i,a) \textnormal{d}c_i\textnormal{d}a
\end{align}
Again, we trained this model using a single objective, using separately resampled subsets $D^a$ and $D^c$ to infer each latent variable (see Supplement). We then tested our model at both one-shot character generation and 5-shot alphabet generation, using samples from previously unseen alphabets. Our single trained model is able to learn structure at both layers of abstraction (Figure \ref{fig:alphabets})

\begin{figure}[h]
\includegraphics[width=0.45\textwidth]{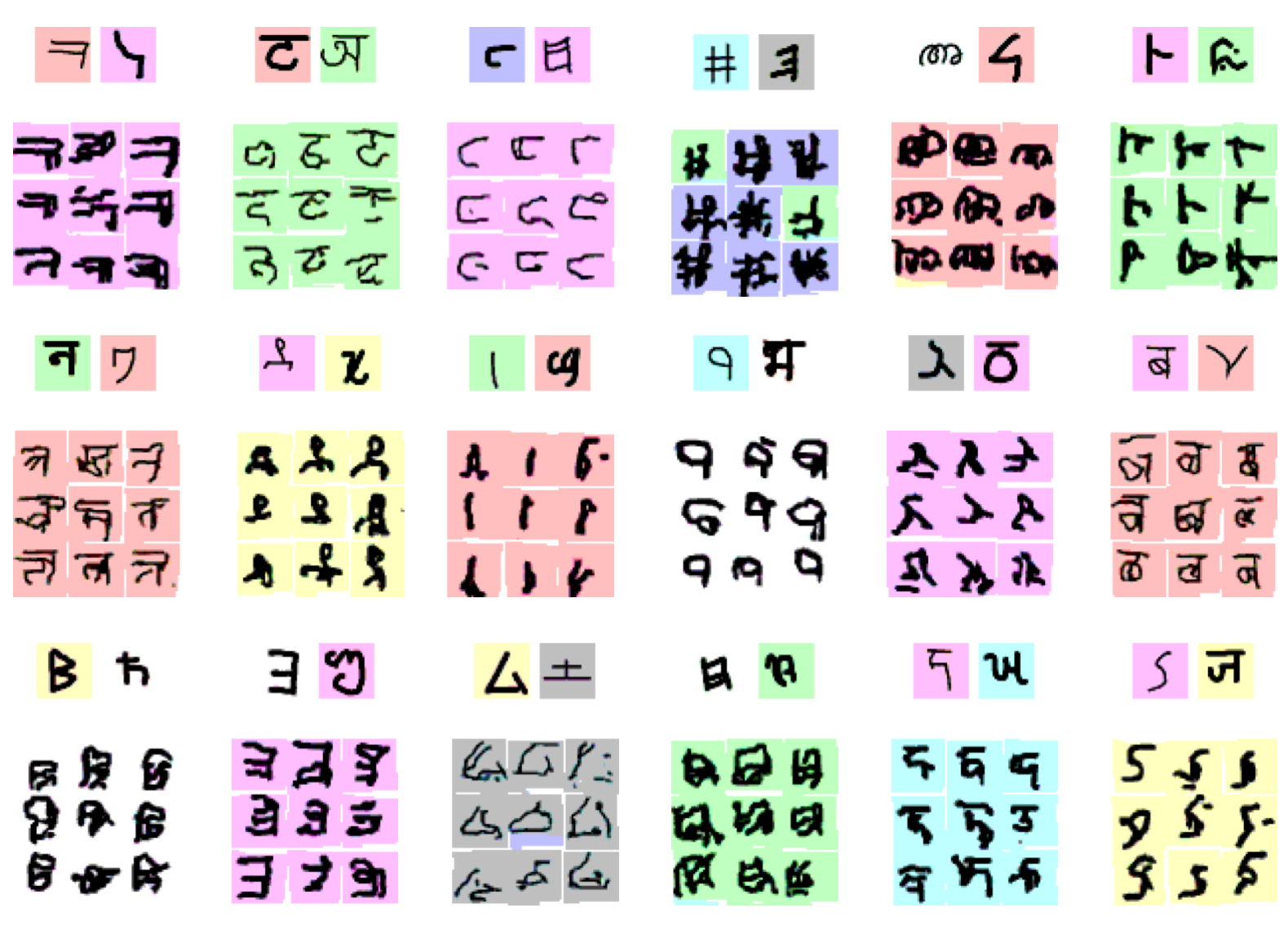}
\caption{Previously unseen characters redrawn with both the colour and stroke width of a second character. For each group, the top two images denote the content (left) and style (right).}
\label{fig:stylecontent}
\end{figure}

For the factorial VHE, we extended the Omniglot dataset by assigning each image to one of 30 randomly generated styles (independent of its character class), modifying both the colour and pen stroke characteristics of each image. We then extended the PixelCNN model to include a 6-dimensional latent variable $s$ to represent the \textit{style} of an image, alongside the existing $c$ to represent the \textit{character}. We used a CNN for style encoder $q(s;D^s)$, and for each image location we condition the PixelCNN decoder using the outer product $s \otimes c_{ij}$.

We then test this model on a \textit{style transfer} task by feeding separate images into the character encoder $q(c;D^c)$ and style encoder $q(s;D^s)$, then rendering a new image from the inferred $(c,s)$ pair. We find that synthesised samples are faithful to the respective character and style of both support images (Figure \ref{fig:stylecontent}), demonstrating the ability of a factorial VHE to successfully disentangle these two image factors using separate latent variables.

\section{CONCLUSION}
We introduce the \textit{Variational Homoencoder}: a hierarchical Bayesian approach to learning expressive generative models from few examples.
We test the VHE by training a hierarchical PixelCNN on the Omniglot dataset, and achieve state-of-the-art results: our model is arguably the first which uses a general purpose architecture to both produce high quality samples and attain near state-of-the-art one-shot classification performance.
We further validate our approach on a dataset of face images, and find that the VHE significantly improves the visual quality and classification accuracy achievable with a PixelCNN decoder. Finally, we show that the VHE framework extends naturally to models with richer latent structure, which we see as a promising direction for future work.

\clearpage
\subsubsection*{References}

\begingroup
\renewcommand{\section}[2]{}%
\bibliography{iclr2018_conference}
\endgroup


\end{document}



\onecolumn
{\Large \bf The Variational Homoencoder:\\ Learning to learn high capacity generative models from few examples\\ SUPPLEMENTARY MATERIAL}
\section{Constrained Posterior Approximation}
In a VAE, use of a recognition network encourages learning of generative models whose structure permits accurate amortised inference. In a VHE, this recognition network takes only a small subsample as input, which additionally encourages that the true posterior $p(c|X)$ can be well approximated from only a few examples of $X$. For a subsample $D \subset X$, $q(c;D)$ is implicitly trained to minimise the KL divergence from this posterior \textit{in expectation} over possible sets $X$ consistent with $D$. For a data distribution $p_d$ we may equivalently describe the VHE objective (Equation \ref{eq:vheobjective}) as
\begin{align}
&\mathop{\mathbb{E}}_{p_d(D)}  \mathop{\mathbb{E}}_{p_d(X|D)} \Bigg[\mathop{\mathbb{E}}_{x \in X}\Big[\textrm{log } p(x)\Big] - \frac{1}{|X|}\textnormal{D}_{KL}\Big[q(c;D) \parallel p(c|X)\Big]\Bigg]
\label{eq:vheobjective2}
\end{align}
Note that the variational gap on the right side of this equation is itself bounded by:
\begin{align}
\label{eq:qobj}
&\mathop{\mathbb{E}}_{p_d(X|D)} \textnormal{D}_{KL}\Big[q(c;D) \parallel p(c|X)\Big] \ge \textnormal{D}_{KL}\Big[q(c;D) \parallel \mathop{\mathbb{E}}_{p_d(X|D)} p(c|X)\Big] \ge 0
\end{align}
The left inequality is tightest when $p(c|X)$ matches $p(c|D)$ well across all $X$ consistent with $D$, and exact only when these are equal. We view this aspect of the VHE loss as regulariser for constrained posterior approximation, encouraging models for which the posterior $p(c|X)$ can be well determined by sampled subsets $D \subset X$. This reflects how we expect the model to be used at test time, and in practice we have found this `loose' bound to perform well in our experiments. In principle, the bound may also be tightened by introducing an auxiliary inference network (see Supplementary Material \ref{auxiliary}) which we leave as a direction for future research.

\section{Tightened variational bound}
\label{auxiliary}
The likelihood lower bound in the VHE objective may also be tightened by introduction of an auxiliary network $r(D; c,X)$, trained to infer which subset $D \subset X$ was used in $q$. This meta-inference approach was introduced in Salimans et al. (2015) to develop stochastic variational posteriors using MCMC inference.
Applied to Equation \ref{eq:vheobjective}, this yields a modified bound for the VHE objective
\begin{align}
\label{eq:tightened}
\textnormal{log }p(\mathcal{X}) &\ge \sum_{x\in \mathcal{X}} \mathop{\mathbb{E}}_{\substack{q'(D;X_{(x)})\\q(c;D)}} \Bigg[ \textrm{log }p(x|c) - \frac{1}{|X_{(x)}|}\textrm{log }\frac{p(c)r(D;c,X_{(x)})}{q'(D;X_{(x)})q(c;D)}\Bigg]
\end{align}

where $q'(D;X)$ describes the stochastic sampling procedure for sampling $D \subset X$, which indeed may itself be learned using policy gradients.

We have conducted preliminary experiments using fixed $q'$ and a simple functional form $r(D;c,X)=\prod_i r(d_i;c,X) \propto \prod_i \big[ f_\psi(c)\cdot \xi_{d_i} \big]$, learning parameters $\psi$ and embeddings $\{ \xi_d: d \in \mathcal{X}\}$; however, on the Omniglot dataset we found no additional benefit over the strictly loose bound (Equation \ref{eq:vheobjective}). We attribute this to the already high similarity between elements of the same Omniglot character class, allowing the approximate posterior $q(c; D)$ to be relatively robust to different choices of $D$. However, we expect that the gain from using such a tightened objective may be much greater for domains with lower intra-class similarity (e.g. natural images), and thus suggest the tightened bound of Equation \ref{eq:tightened} as a direction for future research.

\section{Variational Bound for Hierarchical Models}
\label{hierarchicalbound}
The resampling trick may be applied iteratively, to construct likelihood bounds over hierarchically organised data. Expanding on Equation \ref{eq:vheobjective}, suppose that we have collection of datasets
\begin{align}
\mathbf{X} = \mathcal{X}_1 \sqcup \mathcal{X}_2 \sqcup \ldots \sqcup \mathcal{X}_N
\end{align}

For example, each $\mathcal{X}$ might be a different alphabet whose latent description $a$ generates many character classes $X_i$, and for each of these a corresponding latent $c_i$ is used to generate many images $x_{ij}$. From this perspective, we would like to learn a generative model for alphabets $\mathcal{X}$ of the form
\begin{align}
p(\mathcal{X}) = \int p(a) \prod_{X_i \subset \mathcal{X}} \int p(c | a)\prod_{x \in X_i} p(x|c,a) \textnormal{d}c\textnormal{d}a
\end{align}

Reapplying the same trick as before yields a bound taken over all elements $x$:
\begin{align}
\begin{split}
\textnormal{log }p(\mathbf{X}) &\ge \sum_{x \in \mathbf{X}}\mathop{\mathbb{E}}_{\substack{D^a \subset \mathcal{X}_{(x)} \\ D^c \subset X_{(x)}}} \Bigg[ \mathop{\mathbb{E}}_{\substack{q_a(a|D_1)\\q_c(c|D_2,a)}} \textnormal{log }p(x|c) \nonumber \\ &- \frac{1}{\vert \mathcal{X}_{(x)} \vert} \textnormal{D}_{KL}\big(q_a(a|D^a) \parallel p(a)\big) \\&- \frac{1}{\vert X_{(x)} \vert} \textnormal{D}_{KL}\big(q_c(c|D^c,a) \parallel p(c|a)\big) \Bigg]
\end{split}
\end{align}

This suggests an analogous \textit{hierarchical resampling} procedure: Summing over every element $x$, we can bound the log likelihood of the full hierarchy by resampling subsets $D^c, D^a, $ etc. at each level to construct an approximate posterior. All networks are trained together by this single objective, sampling $x$, $D^a$ and $D_c$ for each gradient step. Note that this procedure need only require passing sampled elements, rather than full classes, into the upper-level encoder $q_a$.

\section{Results for Simple 1D Distributions}
\label{section:simple}
With a Neural Statistician model, under-utilisation of latent variables is expected to pose the greatest difficulty either when $|D|$ is too small, or the inference network $q$ is insufficiently expressive. We demonstrate on simple 1D distributions that a Variational Homoencoder can bring improvements under these circumstances. For this we created five datasets as follows, each containing 100 classes from a particular parametric family, and with 100 elements sampled from each class.

\begin{enumerate}
\item \textbf{Gaussian}: Each class is Gaussian with $\mu$ drawn from a Gaussian hyperprior (fixed $\sigma^2$).
\item \textbf{Mixture of Gaussians}: Each class is an even mixture of two Gaussian distributions with location drawn from a Gaussian hyperprior (fixed $\sigma^2$ and separation).
\item \textbf{von Mises}: Each class is von Mises with $\mu$ drawn from a Uniform hyperprior (fixed $\kappa$).
\item \textbf{Gamma}: Each class is Gamma with fixed $\beta$, and with $\alpha$ drawn from a Uniform hyperprior.
\item \textbf{Discrete}: Each class is Uniform on a subset of \{1,\ldots,8\}, either \textit{1-4}, \textit{5-8}, \textit{odd} or \textit{even}.
\end{enumerate}

For each dataset, we then trained models using a variety of values for $|D|$, restricting the inference network $q(c; D)$ to a simple linear map with Gaussian output. In each case the generative model $p(x|c)$ was set to the correct parametric family, with parameters learned as a linear function of $c$. All models were built in Torch 7 \citep{torch} and optimised using Adam \citep{kingma2013auto} for 200 epochs. To aid optimisation we used an additional 50 epochs for KL annealing, and used training error to select the best parameters from 3 independent training runs.

Our results show that, when $|D|$ is small, the Neural Statistician often places little to no information in $q(c;D)$ (Figure \ref{fig:plot}, top row). Our careful training suggests that this is not an optimisation difficulty, but is core to the objective as in \citet{chen2016variational}. In these cases a VHE better utilises the latent space, leading to improvements in both few-shot generation (by conditional NLL) and classification. Importantly, this is achieved while retaining good likelihood of test-set classes, typically matching or improving upon that achieved by a Neural Statistician (including a standard VAE, corresponding to $|D|=1$).

\begin{figure*}[h]
\center\includegraphics[width=\textwidth]{simple2}
\caption{Comparison of models trained simple 1D distributions using various alternate objectives, including the Neural Statistician. $|D|$ is the number of encoder inputs during training. \textit{Top row}: Mean encoded information $\textnormal{D}_{KL}[q(c;D) \parallel p(c)]$; \textit{Second row}: $|D|$-shot generation loss $-\mathop{\mathbb{E}}_{c \sim q(c;D)} \text{log } p(x'|c)$; \textit{Third row}: $|D|$-shot binary classification error, by minimising conditional NLL \textit{Bottom row}: Joint NLL (per element) of full test set, calculated by importance weighting on 200 samples from $q(c;X)$; }
\label{fig:plot}
\end{figure*}

\newpage
\section{PixelCNN Omniglot Architecture}
\subsection{Methodology}
Our architecture uses a 8x28x28 latent variable $c$, with a full architecture detailed below.  For our classification experiments, we trained 5 models on each of the objectives (VHE, Rescale only, Resample only, NS).  Occasionally we found instability in optimisation, causing sudden large increases in the training objective. When this happened, we halted and restarted training. All models were trained for 100 epochs on 1000 characters from the training set (the remaining 200 have been used as validation data for model selection). Finally, for each objective we selected the parameters achieving the best training error.

Note that we did not optimise or select models based on classification performance, other than through our development of our model's architecture. However, we find that classification performance is well correlated the generative training objective, as can be seen in the full table of results.

We perform classification by calculating the expected conditional likelihood under the variational posterior: $\mathop{\mathbb{E}}_{q(c;D)}p(x|c)$. This is approximated using 20 samples for the outer expectation, and importance sampling with $k=10$ for the inner integral $p(x|c)=\mathop{\mathbb{E}}_{q(t|x)}\frac{p(t)}{q(t|x)}p(x|c,t)$

To evaluate and compare log likelihood, we trained 5 more models with the same architecture, this time on the canonical 30-20 alphabet split of Lake et al. We did not augment our training data. Again, we split the background set into training data (25 alphabets) and validation data (5) but do not use the validation set in training or evaluation for our final results. We estimate the total class log likelihood by importance weighting, using k=20 importance samples of the class latent $c$ and k=10 importance samples of the transformation latent $t$ for each instance.

\newpage
\subsection{Conditional Samples on Omniglot}

\label{conditionalsamples}
\begin{figure*}[h]
\center\includegraphics[width=0.75\textwidth]{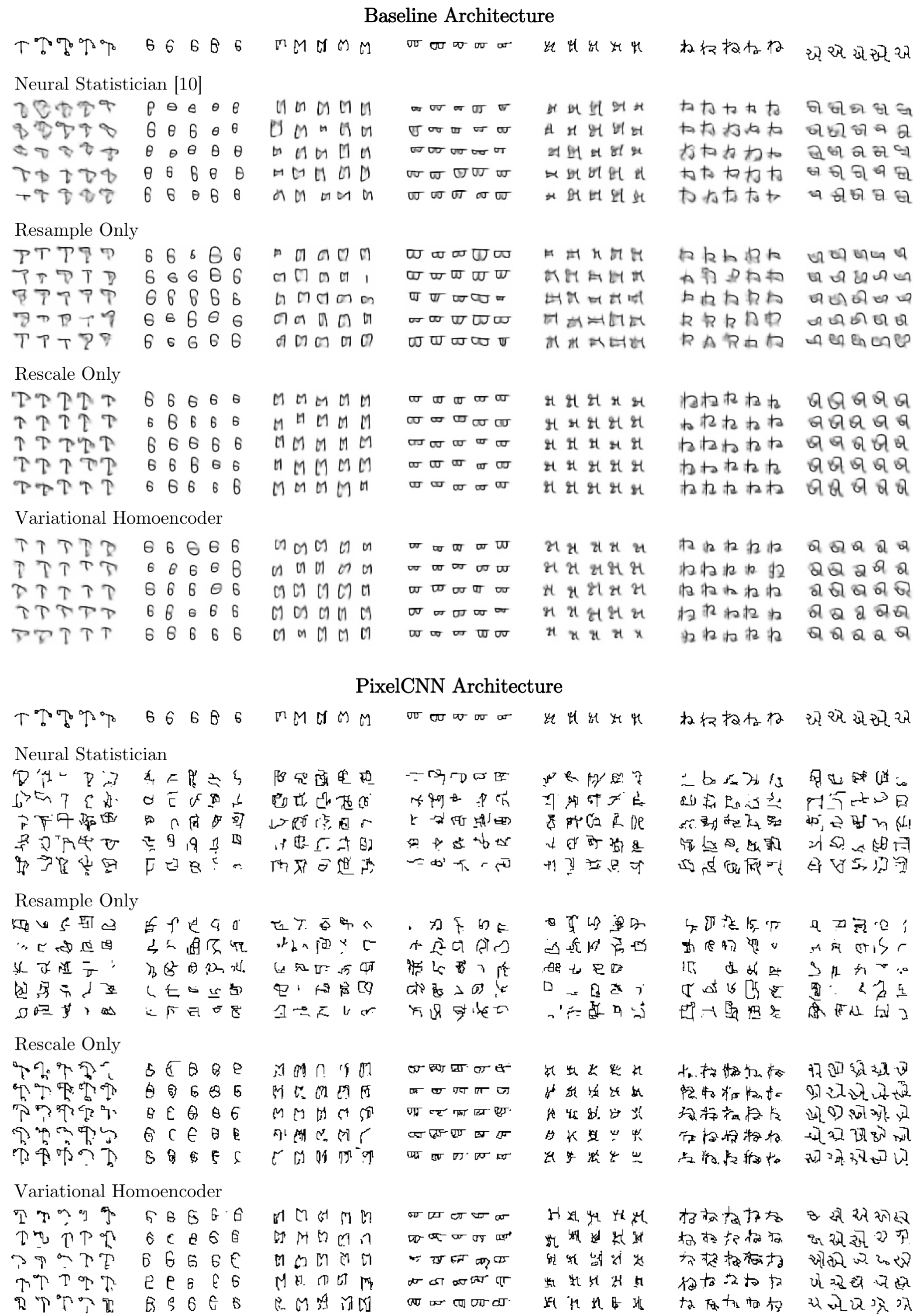}
\label{fig:onepage}
\end{figure*}

\newpage
\subsection{Model Specification}
\label{section:architecture}
[d] denotes a dimension d tensor. \{t\} denotes a set with elements of type t. Posteriors $q$ are Gaussian.
\section*{p(c)}
A PixelCNN with autoregressive weights along only the spatial (not depth) dimensions of c. We use 2 layers of masked 64x3x3 convolutions, followed by a ReLU and two 8x1x1 convolutions corresponding to the mean and log variance of a Gaussian posterior for the following pixel.

\section*{p(t)}
t: [16] ~ Normal(0, 1)

\section*{p(x|c,t)}
c: [8x28x28], t: [16] $\mapsto$ x: [1x28x28]\\
\begin{table*}[H]
\begin{tabular}{lll}
\textbf{Input} & \textbf{Operation} & \textbf{Output}\\
t & Linear & t2: [6] \\
c,t2 & Spatial Transformer & y1: [8x28x28] \\
y1 & 64x3x3 Conv; Relu; BatchNorm & y2: [64x28x28]\\
y2 & 64x3x3 Conv; Relu; BatchNorm & y3: [64x28x28]\\
y3 & 64x3x3 Conv; Relu; BatchNorm & y4: [64x28x28]\\
y4 & 64x3x3 Conv; Relu; BatchNorm & y5: [64x28x28]\\
y5 & 64x3x3 Conv; Relu; BatchNorm & y: [64x28x28]\\
y & PixelCNN & x: [1x28x28]
\end{tabular}\\
\end{table*}
PixelCNN is gated by y, and is autoregressive along only the spatial (not depth) dimensions of c. We use 2 layers of masked 64x3x3 convolutions, followed by a ReLU, a 2x1x1 convolution and a softmax, corresponding to a Bernoulli distribution on the following pixel.

\section*{q(c;D)}
D: \{[1x28x28]\} $\mapsto$ c: [8x28x28]\\
\begin{tabular}{lll}
\textbf{Input} & \textbf{Operation} & \textbf{Output}\\
D & STNq & Y: \{[1x28x28]\}\\
Y & Mean & y: [1x28x28]\\
y & 16x28x28 Conv & mu: [8x28x28], logvar: [8x28x28]
\end{tabular}

\section*{q(t;x)}
x: [1x28x28] $\mapsto$ t: [16]\\
\begin{table*}[H]
\begin{tabular}{lll}
\textbf{Input} & \textbf{Operation} & \textbf{Output}\\
x & 32x3x3 Conv; 2x2 Max Pooling; ReLU; BatchNorm & y1: [32x15x15]\\
y1 & 32x3x3 Conv; 2x2 Max Pooling; ReLU; BatchNorm & y2: [32x8x8]\\
y2 & 32x3x3 Conv; 2x2 Max Pooling; ReLU; BatchNorm & y3: [32x4x4]\\
y3 & 32x3x3 Conv; 2x2 Max Pooling; ReLU; BatchNorm & y4: [32x2x2]\\
y4 & 32x3x3 Conv; 2x2 Max Pooling; ReLU; BatchNorm & y5: [32x1x1]\\
y5 & Linear & mu: [16], logvar: [16]\\
\end{tabular}
\end{table*}[H]

\section*{Spatial Transformer STNq}
x: [1x28x28] $\mapsto$ y: [1x28x28] \\
\begin{table*}[H]
\begin{tabular}{lll}
\textbf{Input} & \textbf{Operation} & \textbf{Output}\\
x & 16x3x3 Conv; 2x2 Max Pooling; ReLU; BatchNorm & y1: [16x15x15]\\
y1 & 16x3x3 Conv; 2x2 Max Pooling; ReLU; BatchNorm & y2: [16x8x8]\\
y2 & 16x3x3 Conv; 2x2 Max Pooling; ReLU; BatchNorm & y3: [16x4x4]\\
y3 & 16x3x3 Conv; 2x2 Max Pooling; ReLU; BatchNorm & y4: [16x2x2]\\
y4 & 16x3x3 Conv; 2x2 Max Pooling; ReLU; BatchNorm & y5: [16x1x1]\\
y5 & Linear & y6: [6]\\
x, y6 & Spatial Transformer Network & y: [1x28x28]\\
\end{tabular}
\end{table*}

\newpage
\section{Hierarchical Omniglot Architecture}

We extend the same architecture described in Appendix B of \citep{edwards2016towards}, with only a simple modification: we introduce a new latent layer containing a 64-dimensional variable $a$, with a Gaussian prior. We give $p(c|a)$ the same functional form as $p(z|c)$, and give $q(a|D^a)$ the same functional form as $q(c;D^c)$ using the shared encoder.

\begin{figure}[h]
\center\includegraphics[width=0.8\textwidth]{alphabetsforluke}
\label{fig:hierarchicalall}
\caption{10-shot alphabet generation samples from the hierarchical model.}
\end{figure}

\newpage
\section{Conditional Samples on Faces Dataset}
\begin{figure*}[h]
\centering
\includegraphics[width=0.8\textwidth]{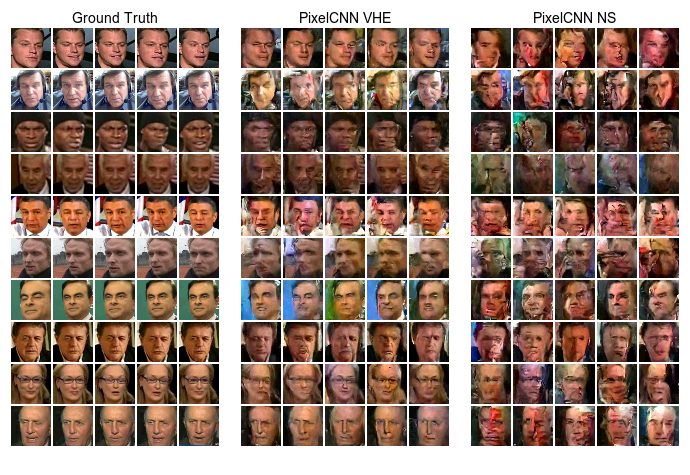}
\caption{5-shot samples of YouTube faces generated using PixelCNN architectures}
\end{figure*}

\newpage
\section{Conditional Samples on Silhouettes Dataset}
We created a VHE using the same deconvolutional architecture as applied to omniglot, and trained it on the Caltech-101 Silhouettes dataset. 10 object classes were held out as test data, which we use to generate both 1-shot and 5-shot conditional samples.
\begin{figure*}[h]
\center
\center\includegraphics[width=0.8\textwidth]{sil}
\label{fig:sil}
\end{figure*}